\documentclass[letterpaper, 10 pt, conference]{ieeeconf}  

\IEEEoverridecommandlockouts                              

\usepackage{graphics} 
\usepackage{epsfig} 
\usepackage{mathptmx} 
\usepackage{times} 
\usepackage{amsmath} 
\usepackage{amssymb}  
\usepackage{mathptmx}
\DeclareMathAlphabet{\mathcal}{OMS}{cmsy}{m}{n}
\usepackage{hyperref}
\usepackage{subfigure}
\usepackage{soul}
\usepackage[ruled, linesnumbered, vlined]{algorithm2e}
\usepackage{comment}
\usepackage{multirow}
\usepackage{graphicx}
\usepackage[table,xcdraw]{xcolor}
\usepackage{colortbl}
\usepackage{notoccite}

\graphicspath{ {./Images/} }

\title{\LARGE \bf UAV-miniUGV Hybrid System for Hidden Area Exploration and Manipulation \vspace{-10pt}}

\author{Durgakant Pushp, Swapnil Kalhapure, Kaushik Das and Lantao Liu 
\thanks{\newline 
D. Pushp, L. Liu are with the Luddy School of Informatics, Computing, and Engineering  at Indiana University, Bloomington, IN 47408, USA. \newline
E-mail:        {\tt\small \{dpushp, lantao\}@iu.edu}. 
\newline
S. Kalhapure is with the Automatic Control and Systems Engineering Department at University of Sheffield, Mappin St, Sheffield. S1 3JD, UK. \newline 
E-mail: {\tt\small sskalhapure1@sheffield.ac.uk}
\newline
K. Das is with the TCS Research, India. \newline
Email:  {\tt\small kaushik.da@tcs.com}
        }%
}

\begin{document}
\maketitle
\thispagestyle{empty}
\pagestyle{empty}
\begin{abstract}
We propose a novel hybrid system (both hardware and software) of an Unmanned Aerial Vehicle (UAV) carrying a miniature Unmanned Ground Vehicle (miniUGV) to perform a complex search and manipulation task. This system leverages heterogeneous robots to accomplish a task that cannot be done using a single robot system. It enables the UAV to explore a hidden space with a narrow opening through which the miniUGV can easily enter and escape. The hidden space is assumed to be navigable for the miniUGV. The miniUGV  uses 
Infrared (IR) sensors and a monocular camera search for an object in the hidden space. The proposed system takes advantage of a wider field of view (fov) of the camera as well as the stochastic nature of the object detection algorithms to guide the miniUGV in the hidden space to find the object. Upon finding the object the miniUGV grabs it using visual servoing and then returns back to its start point from where the UAV retracts it back and transports the object to a safe place. In case there is no object found in the hidden space, UAV continues the aerial search. The tethered miniUGV gives the UAV an ability to act beyond its reach and perform a search and manipulation task which was not possible before for any of the robots individually. The system has a wide range of applications and we have demonstrated its feasibility through repetitive experiments. \\
\end{abstract}

\section*{Supplementary Material}
Video: \url{https://youtu.be/LlW-ylcU0mk}

\vspace{8pt}
\section{Introduction}

\begin{figure*}[ht!] 
  \centering
  \subfigure[]
  	{\includegraphics[height=1.1in, width=1.66in]{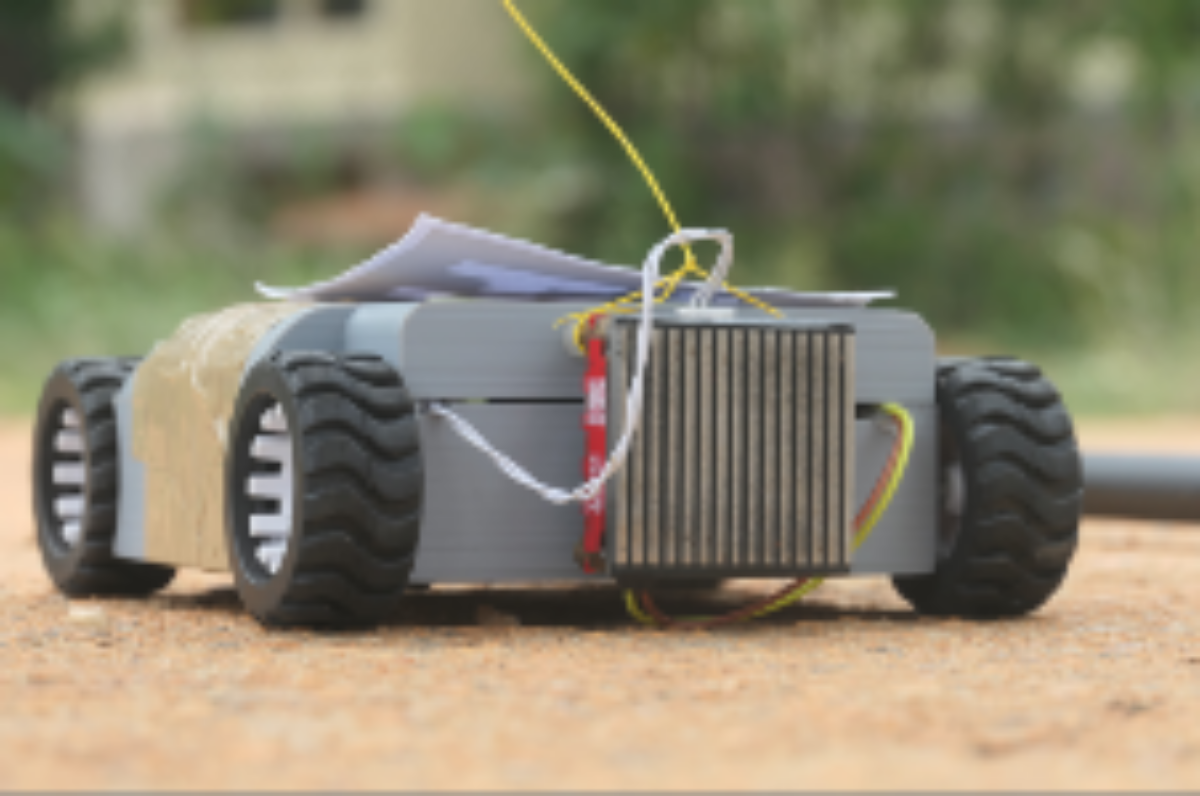}
         \label{fig:title_a}}
  \subfigure[]
  	{\includegraphics[height=1.1in, width=1.66in]{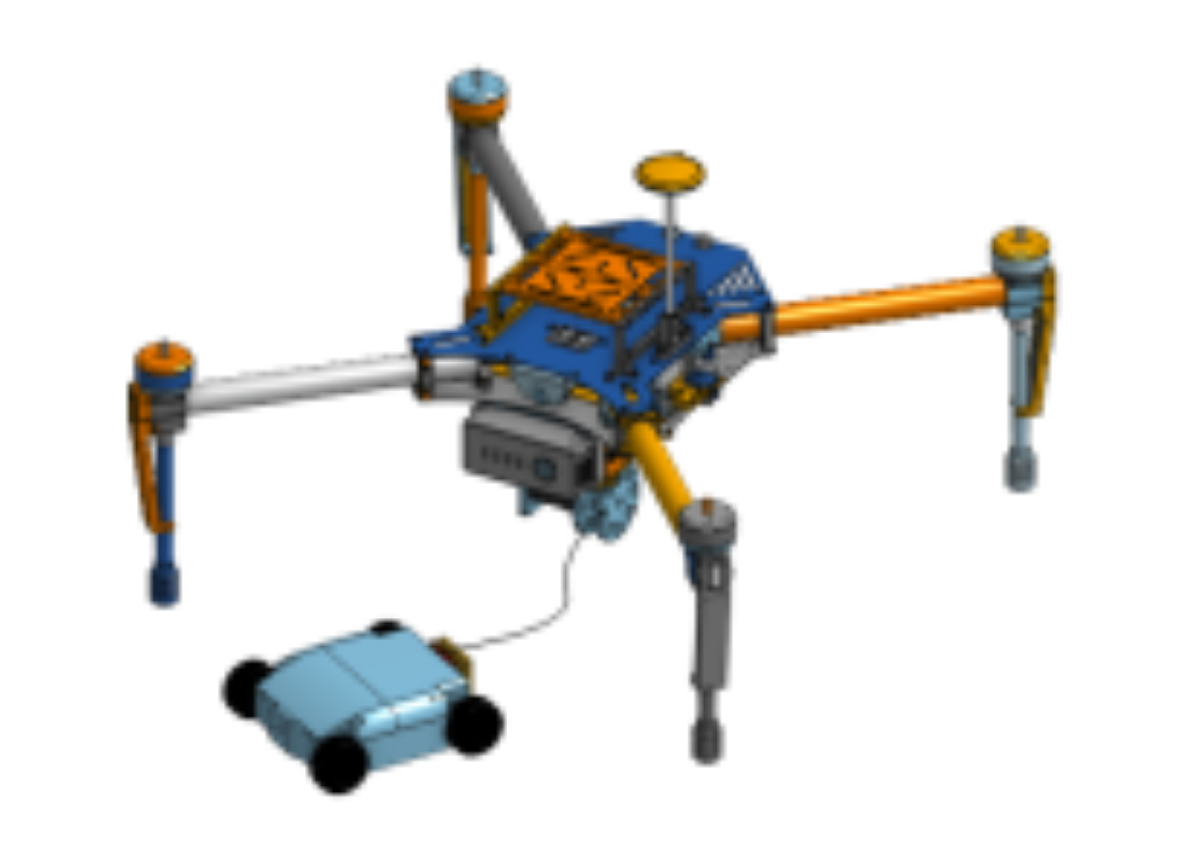}
         \label{fig:title_b}}
  \subfigure[]
  	{\includegraphics[height=1.1in, width=1.66in]{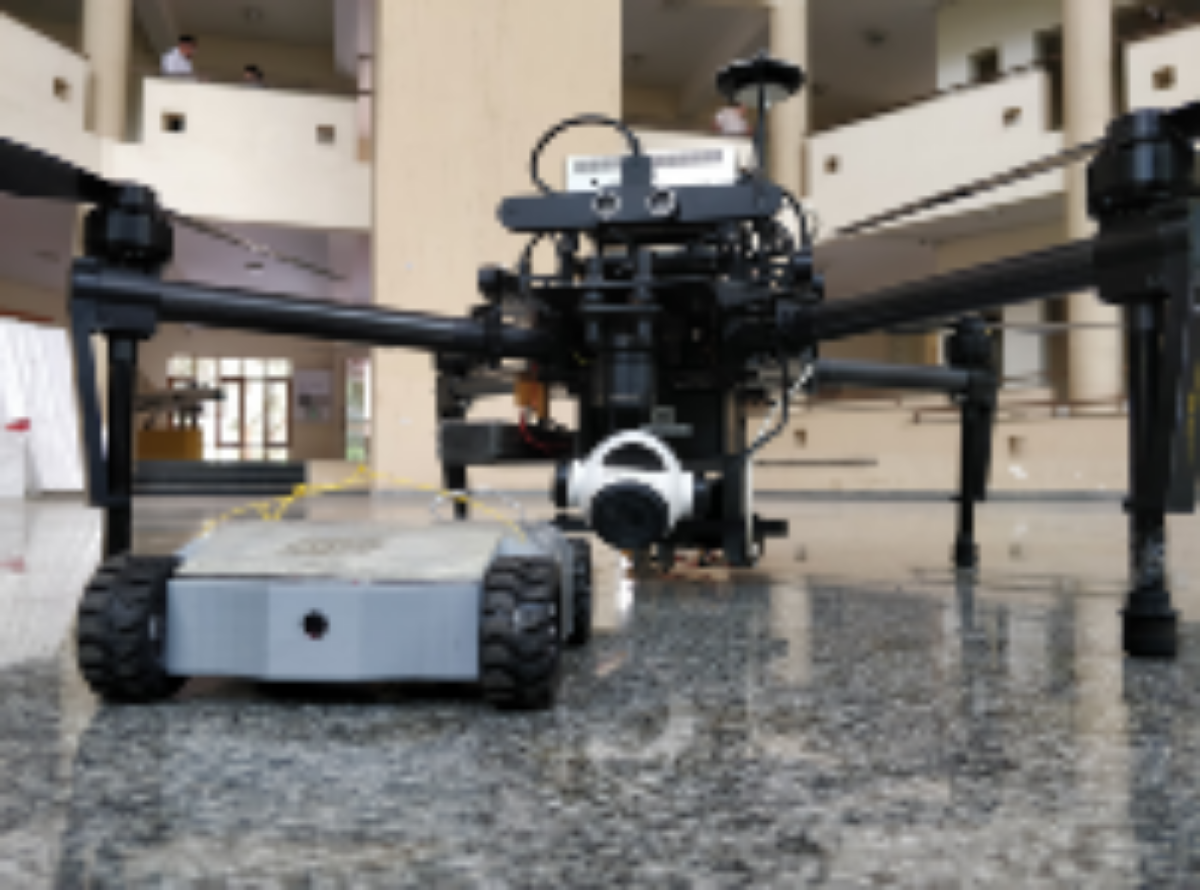}
         \label{fig:title_c}}
  \subfigure[]
  	{\includegraphics[height=1.1in, width=1.66in]{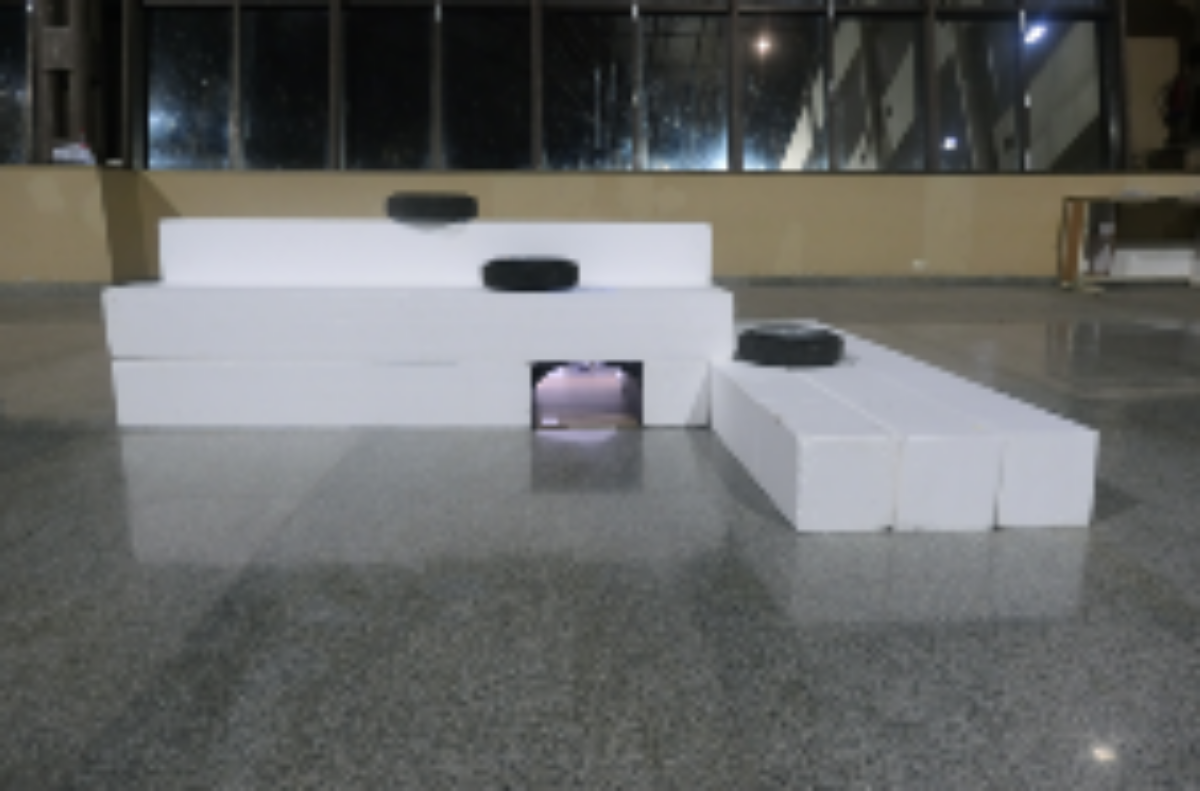}
         \label{fig:title_d}}
  \caption{Proposed UAV-miniUGV hybrid system. (a) shows the developed miniUGV with an EPM at the back and a monocular camera at the front. (b) shows the CAD model of the complete system. (c) shows the real system. (d) shows the replication of a hidden space in the lab. We use this hidden space to understand the behavior of the developed methods in the real world.  
  }
\label{fig:hybrid-system}  
\end{figure*}

To determine the fitness of anything made and to guarantee a safe workplace, a careful examination is required to identify and record hazards for corrective actions. Therefore, a variety of autonomous robots have been developed by researchers and the industry 
{for applications such as area surveillance \cite{Stolfi2021, MISHRA20201, Ganesh2015}, generating detailed maps \cite{Qin2019}, performing complex manipulation tasks \cite{aerial_grasping}, \cite{Srivastava_2021} and retrieval tasks \cite{Shankar2018, Corke2004}, which require robots to carry objects, deploy sensors, avoid obstacles, move in steep and rough terrains. }
For a long time, UAVs have been a center of attraction for inspection and exploration tasks because of their ability of rapid areal scanning. For the scenarios like Search and Rescue (SAR) or natural disasters, the information collected by the flying sensors is not sufficient. In some cases, we specifically want to look inside the areas which remain hidden from the UAV's sensors. We call this space a {\em hidden space} and in this work, we propose a hardware architecture equipped with an autonomy software system to collect information from hidden space as well as to perform some manipulation tasks inside, like eliminating a potential threat from that space.

Our aim {is to} design a hybrid system that leverages the strengths of heterogeneous autonomous platforms in order to accomplish a complex mission. Fig. \ref{fig:hybrid-system} shows the proposed hybrid system that can efficiently perform hidden region exploration and manipulation. 
In greater detail, let us assume that a UAV has scanned a given area and selected the locations where deep scanning is required. By deep scanning, we mean there exists a hidden space that cannot be explored using a UAV but may be an ideal navigable space for other types of robots, for instance, a ground robot or a quadruped. 
The hidden area inspection clearly demands a new hybrid system that must use a UAV as the {\em Primary Robot (PR)} despite having limitations on maximum payload and total flight time. Our proposed system uses a UAV as the PR and a UGV as the {\em Secondary Robot (SR)}. 
In this paper, we provide our system design details for both hardware and software to achieve the required autonomy capability. 
The primary technical challenges identified to make this system function in the real world are as follows: 
\begin{itemize}
  \item Design a UAV-UGV system physically coupled using a passive tether; 
  \item Design a miniature robot with autonomous navigation, exploration and manipulation capabilities, even though the miniature robot is equipped only with limited low-end sensors and computation board;
  \item Design a gripper that can be installed on {the UGV} to pick and place objects in a hidden space;
  \item Design a robust software-system architecture to support repetitive operations.
\end{itemize}

\subsection{Our Contribution}
\label{contribution}
Since the entire architecture involves numerous components, there are a lot of challenges to overcome before being fully autonomous in any real-life scenarios. All of these challenges cannot be addressed in one attempt. Therefore, in this work, we mainly focus on the overall functionality of the proposed hardware and software system to accomplish the hidden object searching and retrieving mission. 
We have designed a miniUGV under the constraints imposed by the UAV, capable of autonomous exploration and manipulation. 
{We are using a tether-controlled mechanism on the UAV to retract and release the miniUGV.} 
An Electropermanent Magnet (EPM) has been installed on the miniUGV to grab magnetic materials if it finds necessary to take that object out of the hidden space during an inspection. A different type of gripper can be designed and installed on the miniUGV to grab non-magnetic materials. We have used EPM to prove the concept that the miniUGV can also be used as a manipulator. 
To be explicit, the main contributions are 
\begin{itemize}
  \item We propose a hybrid system with a UAV as the PR and a miniUGV as the SR with autonomy software architecture for area inspection and manipulation. 
  The architecture supports periodical area inspection where the UAV performs aerial inspection and it uses miniUGV to inspect the hidden space. 
  \item Our in-house made miniUGV is equipped with autonomous navigation and visual servoing capabilities that can easily escape through a $15 cm$ wide and $7 cm$ high narrow entrance. With an EPM at the back, the miniUGV can also act as a manipulator to grab light-weight magnetic objects.
  \item To accomplish the mission, we also designed a curiosity-driven exploration heuristics for miniUGV to find a target object in a hidden space using minimal sensing resources i.e., IR sensors and camera.
 \item  
 {Finally, the hardware schematics, architecture design and the software code for the proposed system is made available on the public domain.} 
\end{itemize}

\section{Related Work}
Recent advancement in embedded platforms has equipped unmanned robots with low form-factor and high computational capability, leading to an increase in the number of autonomous applications. For fast area search and mapping, the UAV is an ideal robot but faces difficulty to maneuver in a constrained environment, whereas the UGV has a longer duration of service, can carry a large battery, better computation devices, and sensors. Over the decades, their cooperative efforts are seen in tasks such as landing of UAV over UGV \cite{landing1}, precision agriculture \cite{agri1}, area mapping where a UAV searches the space and safely lands on UGV for long-duration travel \cite{area_map2}, path discovery and planning \cite{pathplan1}, civil infrastructure inspection \cite{inspect_app3}, etc. 
In \cite{Papachristos2014}, the authors employed a UGV-UAV system to use a powered tether mechanism, which allows heavy batteries to be placed on the UGV and carry power via cables to the UAV platform, in order to achieve longer flight time. Further, a combined path planning strategy is proposed that considers the UAV’s tether constraint kinematics along with the UGV kinematics to find an optimal path in a map based on RRT* algorithm. A passive tether like a rope between the UGV and UAV is used \cite{Miki2019}, with an objective to enable a UGV platform to climb a vertical cliff. 
A similar study by \cite{Shankar2018}, uses tethered hooks attached to UAV platforms to re-capture parachute deployed sensor nodes for surveillance.
Scenarios of high-rise building’s spray painting where drone sprays the paint and UGV powers the drone \cite{tether_app3}, as well as cleaning glass and facade where UGV rolls on rails to support the UGV-UAV cleaning system~\cite{tether_app2}, have helped in reducing the human life risk. Although in these studies, researchers developed mechanisms or techniques to physically connect UAV and UGV together, they generally only focus on considering UAV as a secondary platform.
In contrast, we used UAV as a primary platform that makes it move independently to reach a target location inaccessible to UGVs.

Also related is the infrastructure inspection which requires sensors transported to the test site with teams of trained inspectors. With the help of UAVs, this task can be deployed more frequently and consistently than human operators, thus provides a huge decrease in inspection cost (time, effort, manpower, repair expense, life risk involved), covers a large area, and can be used in hazardous situations. The UAV provides a quick visual scan of infrastructure while the UGV is required to access the narrow regions which are not accessible to the UAV. This collaborative UAV-UGV team has the potential to deliver higher fidelity data and a better scanning coverage than either platform could provide independently.
In \cite{inspect_app3}, an approach for civil inspection has a permanent UGV housing mounted in UAV's landing gear. 
{To involve UGV in the mission, the UAV has to land, thus a pick and drop mechanism is required to increase the effective operation time of the secondary robot. In contrast, we physically couple UGV with UAV using a passive tether that not only makes it reusable but also enables the UGV to act as an extended manipulator.}

To organize the paper, Section \ref{system_overview} explains the system overview, and Section \ref{hardware} provides the hardware design including realization details of a miniUGV system. Section \ref{CDOS} discusses the autonomy software architecture and methods 
using IR sensors and cameras as main onboard sensors. Section \ref{experimental_results} describes the experimental setup and results followed by a conclusion in Section \ref{conclusion}.

\section{System Architecture Overview}
\label{system_overview}

\begin{figure}[t!]
\centering
\includegraphics[width=0.98\linewidth]{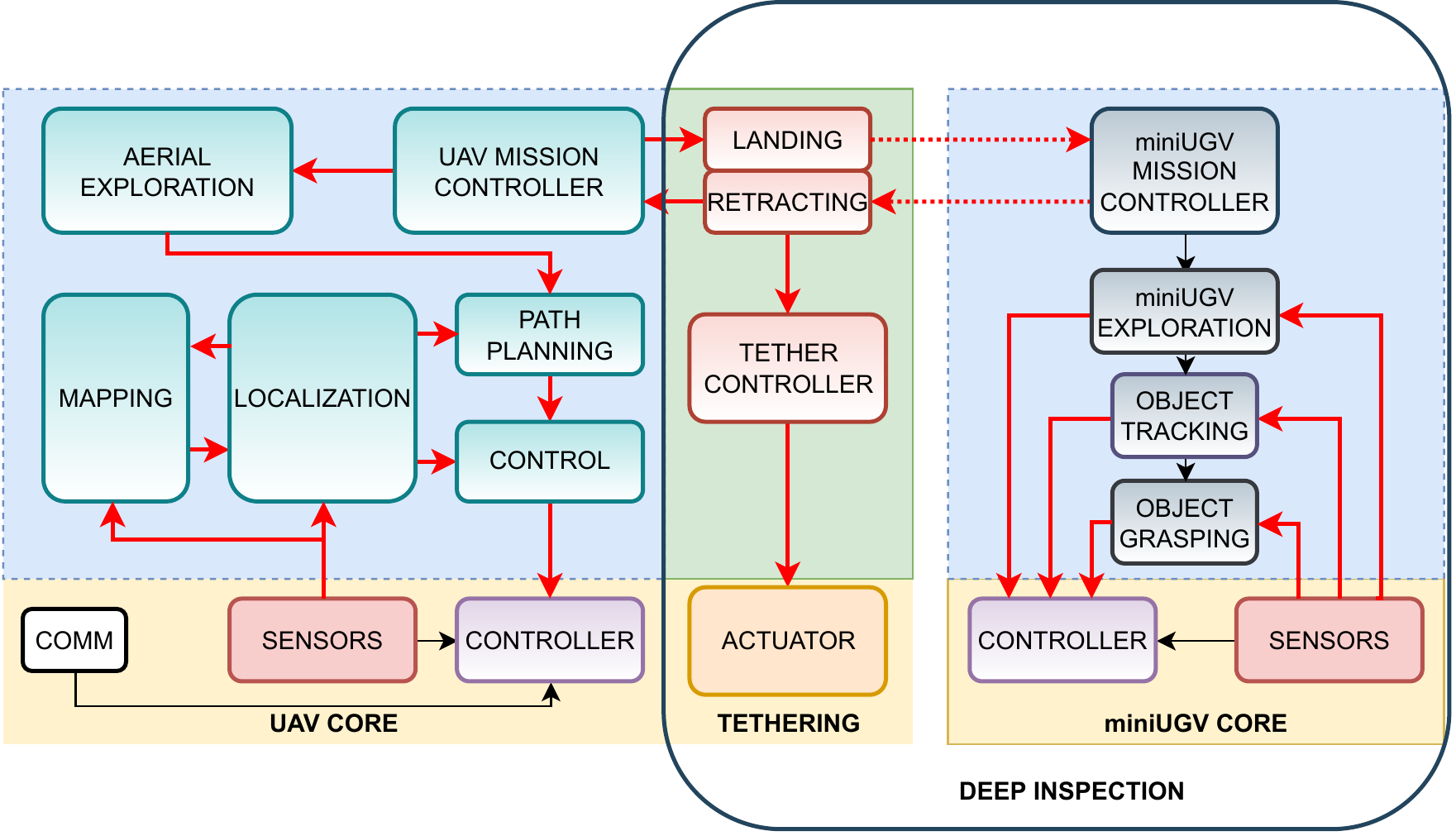}
\caption{An overview of the system architecture. Red dotted lines represent the wireless communication and red solid lines represent wired communication. COMM represents the radio signal transmitter module which is used to take manual control of the mission in case of emergency. 
\vspace{-0.5cm}
}
\label{architecture}
\end{figure}

As shown in Fig. \ref{architecture}, the proposed hybrid system consists of a UAV, a miniUGV, and a tether-controller to retract and release the miniUGV at the entry point of a hidden space. The UAV Mission Controller module (UAVMC) triggers the Aerial Exploration, and the UAV being the PR starts a search operation. Simultaneously, a hidden space detection algorithm runs inside the UAVMC. Identifying the hidden space is an interesting and challenging problem and it needs to be addressed separately. In this work, we assume that a visual marker \cite{marker} is present at the opening of each hidden space. Once a hidden space is detected, UAVMC stops the exploration and triggers the Landing module which detects the landing zone and starts releasing miniUGV after aligning the UAV over the landing point.
The Inertial Measurement Unit (IMU) installed on the miniUGV continuously gives feedback during the landing process. Fig. \ref{fig:Tether_agv} shows that there are two tether attachment points on the miniUGV and it is selected in a way that it always prevents a vertical touch down angle to make sure a successful landing. 

\subsection{Mode Switching}
After performing a successful landing, the miniUGV performs two operations - hidden aerial exploration and object grasping. The UAVMC takes the decision for switching the mode from aerial to ground operations based on the released string length while the sequential decisions to change the mode from exploration to object tracking and grabbing is performed by the miniUGV Mission Controller (miniUGVMC). 

\subsubsection{Aerial Operations}
To explain the mode switching, let us assume that $x=[x_\mathcal{W}, y_\mathcal{W}]$ be the pose of miniUGV, where $x_\mathcal{W}$ and $y_\mathcal{W}$ are the position coordinates in world frame $\mathcal{W}$. 
Let $\mathcal{X}=[\mathcal{X}_\mathcal{W}, \mathcal{Y}_\mathcal{W}, \mathcal{Z}_\mathcal{W}]$ be the pose of UAV in $\mathcal{W}$, where $\mathcal{Z}_\mathcal{W}$ represents the height of center of mass (COM) of the UAV. Let $L$ be the total string length, $l_0$ is the distance of entry point of hidden space from the COM of the UAV, and $l_t$ is the current length of the released string. The ground operations are constrained by the string length and therefore, we use $l_t$ to switch modes as $l_t= 0$ for aerial exploration, $l_t \in (0, \mathcal{Z}_\mathcal{W}]$ for miniUGV landing and, $l_t \in [l_o, L]$ for hidden area exploration. The miniUGV lands near to the entry point of hidden space that allows us to assumed $(l_o - \mathcal{Z}_\mathcal{W}) \approx 0$. 

\subsubsection{Ground Operations}
The SR performs three tasks in deep exploration mode - exploration to find a target object, object tracking, and object grasping. The miniUGVMC triggers a sequential decision making process by initiating hidden area exploration. If the target object is discovered inside the hidden space, Object Tracking (OT) gets triggered. The OT uses visual servoing to reach to the target object and a pre-defined grabbing maneuver is performed to grab the object using EPM.

{Considering the low computational capability of miniUGV, we designed and tested an exploration algorithm to make the system functional in the real world.} 
The proposed system design along with the code, CAD files and PCB design are open-sourced\footnote{ 
{https://github.com/scifiswapnil/UAV-uGV-system}}.

\begin{figure}[t!] 
  \subfigure[miniUGV touch down angle]{\label{fig:Tether_agv}\includegraphics[width=0.46\linewidth]{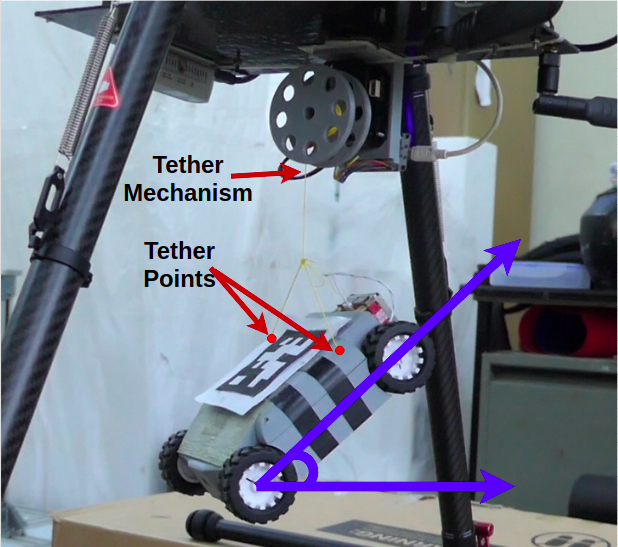}}\quad
  \subfigure[Sensor Coverage]{\label{fig:sensor_cov}\includegraphics[width=0.48 \linewidth, height=0.405\linewidth]{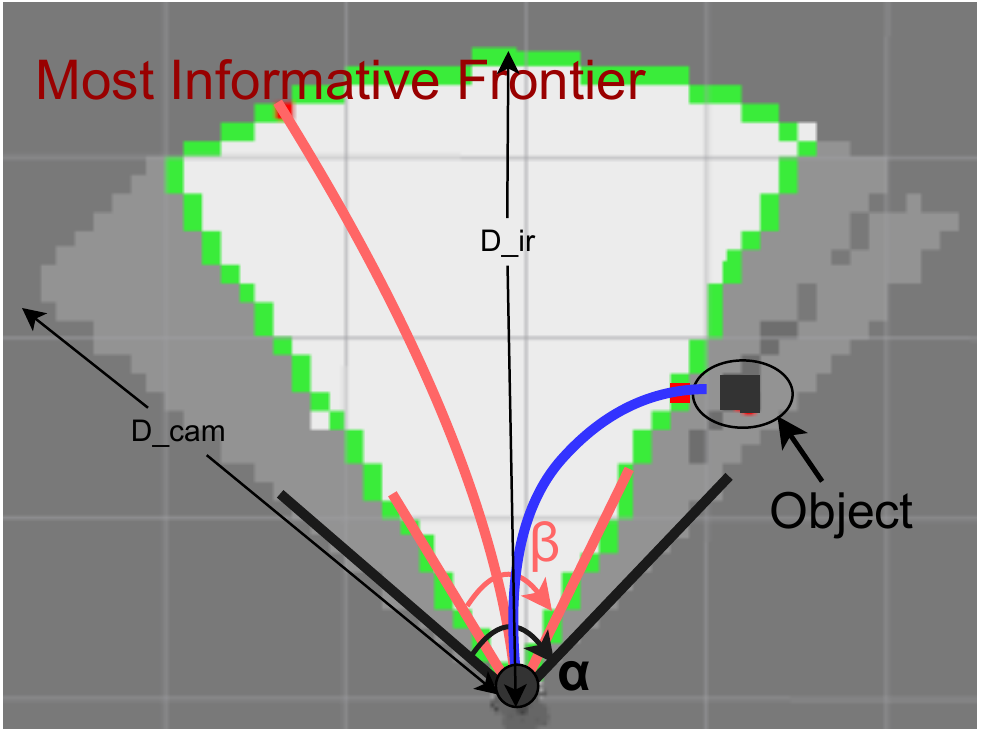}}
\caption{(a) Shows the tethering mechanism and (b) shows Dual-Sensor coverage. Green-colored grids are frontiers in IR's field of view. Red grids are the two frontiers selected on different parameters.
\vspace{-15pt}}
\label{fig:Tether mechanism}
\end{figure}

\section{Hardware System Realization}
\label{hardware}
The choice of SR highly depends on the type of hidden space. For instance, a miniUGV is the best suitable SR for exploring the narrow and closed space shown in Fig.~\ref{fig:hybrid-system}.
Tether-controlled release and retract system can also be replaced with a pick and drop mechanism to make the SR move freely inside the hidden space. Removing a physical connection between the two robots can have a huge advantage in the exploration of a more complex hidden space with obstacles but at the same time, picking the SR becomes more challenging. Hence, we opt for the tether-controlled pick and place of the SR in this work. 

\subsubsection{miniUGV Design}
\begin{figure}[t!] \vspace{6pt}
\centering
\includegraphics[width=0.8\linewidth, height=1.7in]{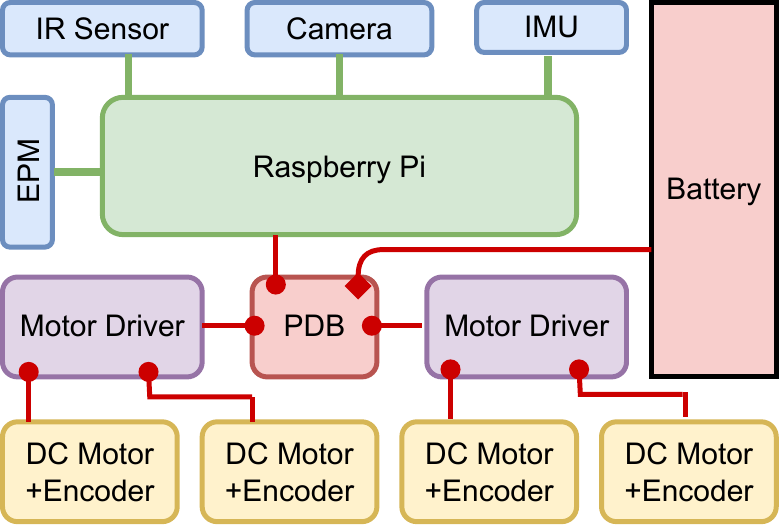}\quad
\caption{Hardware overview of our custom-made miniUGV. Power Distribution Board (PDB) distributes the power from battery to all the components.}
\label{fig:ugv_arch}
\end{figure}

We have developed a custom-made miniUGV platform with onboard computing for path planning and vision processing. The platform is a 4-wheel drive mobile robot with a footprint of  $130~mm \times  120~mm \times  55~mm$. The chassis of the mobile robot is designed to house all the hardware components inside and manufactured using an additive manufacturing technique. The chassis is manufactured using PLA (Poly-lactic Acid) filament that is durable and lightweight material for the mobile platform. Fig.\ref{fig:ugv_arch} shows the hardware architecture diagram of the miniUGV with various components and their interfaces.

\begin{figure}[t!] 
{
\centering
\includegraphics[scale=0.15]{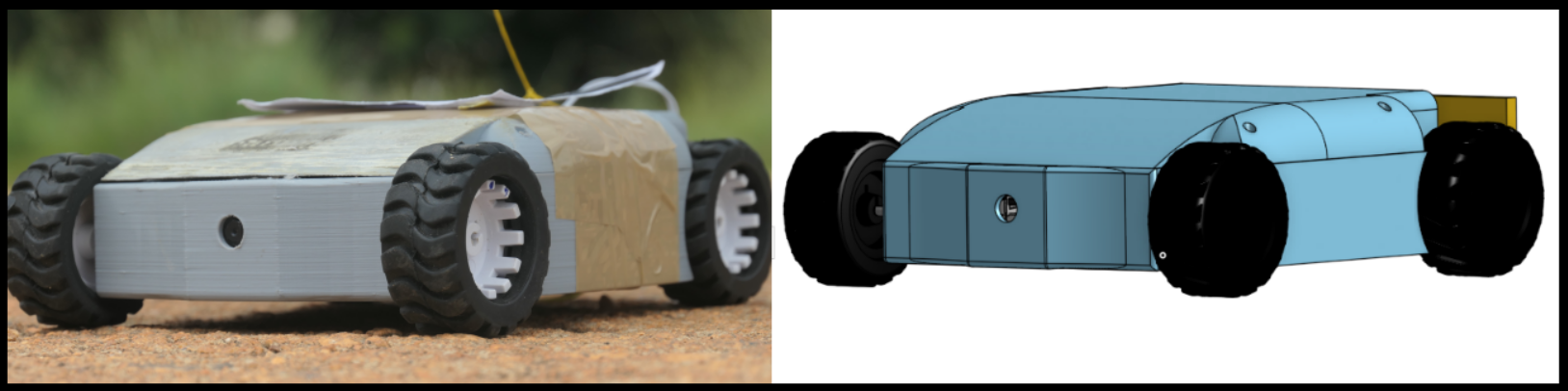} 
\caption{The miniUGV platform (Left: real robot, Right: CAD model) \vspace{-10pt}
}
\label{fig:mini-UGV}
} \vspace{-10pt}
\end{figure}

The onboard compute unit is raspberry pi 3 Model B+, with an 8-megapixel camera and a peripheral interface for 9 degrees of freedom IMU, closed-loop motor drive control, and IR obstacle sensor. The DC motors with quadrature hall-effect encoders are used and controlled using the two H-bridge driver boards and the quadrature encoder feedback to the compute unit. The motor drives have a closed-loop PID controller for velocity and position control of the motors. The IMU is interfaced with the compute unit via I2C, which gives 3 axes acceleration, 3 axes gyro rate, and 3 axes magnetometer data, for estimating the orientation of the miniUGV in 3D space. The IMU unit houses an onboard digital motion processor that performs independent step counting, quaternion calculation, and orientation estimation, effectively removing the communication delays and offloading the computation task off the compute unit. The platform is equipped with an electro-permanent magnet (EPM) by OpenGrab, where the external magnetic field can be controlled using an electric pulse around a permanent magnet acting as a magnetic gripper. The miniUGV platform is powered by a 2S 1000mAh lithium-polymer battery with a battery management system (BMS) that monitors and controls the charge/discharge rate. Effectively, the miniUGV can operate for a {\em continuous 18 minutes}, where the processor is running at 95\% capacity and all the sensors and drive motors are actively operational. Fig.\ref{fig:mini-UGV} shows the CAD model of the miniUGV and the manufactured unit. The miniUGV is connected via a tether to the UAV platform. For communication, the miniUGV uses the raspberry pi's onboard WiFi network.  

\subsubsection{Object Grabbing}
The miniUGV is equipped with an EPM for object grabbing. When the miniUGV reaches close to the target object after a particular minimum distance d\textsubscript{min}, it rotates a full 180 degrees in place that points the EPM to face the target object. Then the EPM is triggered and, the miniUGV
moves backward to push the object. This push confirms the contact between the EPM and the object.

\subsubsection{Tether mechanism}
 The tethering module is a setup responsible for maintaining the tether distance between the UAV and the miniUGV. The tether mechanism is made of a Robotis-Dynamixel MX-38 motor. This motor is interfaced via a driver module which is connected to a micro-controller that drives it at a particular velocity to a position. The Dynamixel motor has a rotatory spindle that unwinds the tether with a certain velocity based on the tether required or winds the tether to retract the miniUGV. 
 The miniUGV communicates its current pose, desired velocity and desired acceleration information to the tethering module over WiFi.
 The tether controller is a feedback control loop implemented on the micro-controller unit that continuously tracks the miniUGV's and the UAV's positions, computing the tether distance requirement to drive the rotor.
 
\textit{Controller:} Based on the miniUGV's position feedback, a taut control loop \cite{Nicotra2014}  either releases or retracts the tether to maintain a positive tension on the cable. Inspired by Tal Glick {\em et al's} work \cite{Glick2018} on an unified model approach for tether control, a model predictive control (MPC) loop is developed for the taut control of the tether. In \cite{Glick2018} and \cite{Lee2015}, the UAV platform is connected via a tether cable to a ground station. In our proposed system, the drone is considered as a stable (position-controlled) platform having a winching mechanism that controls the length of the tether. The miniUGV on the other hand has fixed knot. The MPC controller takes input as the desired velocity and acceleration, based on which the winching mechanism either releases or retracts the tether.

\section{Autonomy Software System}
\label{CDOS}

To search an object in an unknown environment, a robot needs to first know where it is located in that environment. This can be achieved through the matured simultaneous localization and mapping (SLAM) technique~\cite{slam}.
In addition to that, since our task is the object searching with onboard sensors, we will also need to perform state estimation for the target object, and further drive the miniUGV to quickly discover and fetch the object. 
Fig.\ref{fig:sensor_cov} shows that the miniUGV is equipped with two types of sensors - IR sensors of maximum sensing range $d_{ir}$ and a field of view (fov) $\beta$, and a monocular camera of fov $\alpha$. 
{Typically, $\beta$ is much smaller than $\alpha$. If the miniUGV relies on IR sensors for space exploration and uses a camera only for object detection, it takes more time to find the object. In this section, we will discuss a way to include camera observations in the exploration process.}  

\subsection{{Object Occupancy Mapping with Camera}}
\label{sect:cam_model}
We estimate the state of the object through a probabilistic way. 
Similar to the design of {\em occupancy grids} \cite{PRbook}, we assume another occupancy map $\mathcal{O} = \{{{\omicron}}\}$ that stores the probabilities of grid cells being occupied with any given object of interest (instead of the whole environment). Let us call it an {\em object map}. In order to find the probability of occupancy of a given object in $\mathcal{O}$, we assume that the object detection algorithm provides the detection status as $true$ or $false$ along with a confidence value. We consider this confidence as the probability of occupancy of a grid with the given object. 
{However, we must find the distance of the detected object in the camera frame.} The confidence in object detection is the highest when all the features describing that object are clearly visible. The visibility decreases with the increasing distance of the object from the camera. Hence, for any image based object detection algorithm, we assume that
the confidence value $conf = \eta/\left| d \right|$, where 
$d$ is the distance of the object in the camera frame and $\eta$ is the constant of proportionality.   
Let $\mathcal{D}_{cam}$ be the maximum distance an object can be detected by an object detection algorithm (e.g., \cite{Anand2019}). With that, we assume a monocular camera acts like a 2D range finder with a fov $\alpha$ (same as the fov of the camera) and with a maximum sensing range $\mathcal{D}_{cam}$, that measures the probability of occupancy of the object in $\mathcal{O}$. 
The distance of an object of known size can be obtained as
$d = \frac{k_c}{\gamma}$
where $k_c$ is an intrinsic constant of the camera and $\gamma$ is the pixel per meter (ppm) value. The ppm is defined as the length of an object in an image divided by the length of the object in the real world \cite{Anand2019iros}. 

Let $\zeta_{1:t}$ be the set of all measurements up to time $t$, and $x_{1:t}$ is the path of the miniUGV defined as the sequence of all poses. 
The probability that a grid $i$ in $O$ is occupied with the given object is denoted as $p(\omicron_i|\zeta_{1:t}, x_{1:t})$ which is updated incrementally in the fashion of a Bayesian filter \cite{PRbook}, 
\begin{equation}
\label{p_o_update}
p(\omicron_i|\zeta_{1:t}, x_{1:t}) = \frac{p(\omicron_i|\zeta_t, x_t) p(\zeta_t|x_t) p(\omicron_i|\zeta_{1:t-1}, x_{1:t-1})}{p(\omicron_i) p(\zeta_t| \zeta_{1:t-1}, x_{1:t}) },
\end{equation}
where $p(\omicron_i|\zeta_{1:t-1}, x_{1:t-1})$ represents the occupancy probability in the previous time step, and prior $p(\omicron_i)$ is set with probability 0.5 in the initial step. 

Furthermore, the grids in $\mathcal{O}$ 
can be represented as 
    $\mathcal{O} = \mathcal{O}_{free} \cup  \mathcal{O}_{occ} \cup \mathcal{O}_{unk}$, 
where $\mathcal{O}_{free}$, $\mathcal{O}_{occ}$ and $\mathcal{O}_{unk}$ are the set of $free$, $occupied$, and $unknown$ grid cells respectively. 
In practice we determine the occupancy probabilities as follows, 
\begin{equation}
\label{cap_O}
    p(\omicron_i |\zeta_{1:t}, x_{1:t})= 
\begin{cases}
    0,[Free]& \text{if } p(\omicron_i|\zeta_{1:t}, x_{1:t}) \in (0, \lambda_1)\\
    0.5,[Unk]& \text{if } p(\omicron_i|\zeta_{1:t}, x_{1:t}) \in [\lambda_1, \lambda_2]\\
    p(\omicron_i |\zeta_{1:t}, x_{1:t}),& \text{if } p(\omicron_i|\zeta_{1:t}, x_{1:t}) \in (\lambda_2, 1)\\
\end{cases}
\end{equation}
Here, $\lambda_1, \lambda_2 \in [0, 1]$ are the confidence in object detection. (In our experiments we empirically set $\lambda_1=0.10$ and $\lambda_2=0.95$.)

\begin{figure*}[t]  
{
\centering
\includegraphics[scale=0.56]{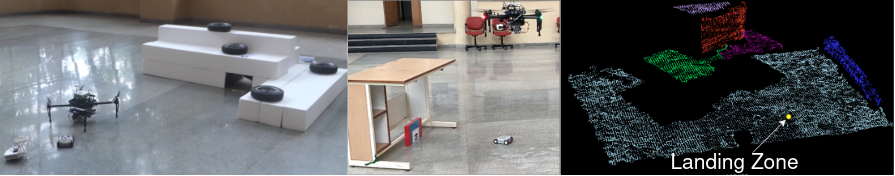} 
\vspace{-0.6cm}
\caption{Experimental setup. First and second images from the left shows snapshots of the real experiment conducted for two different types of hidden spaces with the proposed system. The last image is the detected landing zone.  \vspace{-8pt}}
\label{fig:exp-images}
}
\end{figure*}

\subsection{Occupancy Grid Mapping with IR Sensors}
Since compared to the camera the IR has a shorter range and a limited fov, we create a different occupancy grid map with only the IR perception. This shall help define the curiosity-based exploration which will be discussed later. 
Specifically, we approximate the hidden space $h$ with $N$ number of 2D grids  $G = \{{{g}}\}$, where each grid $g_i$ is represented by its center. The 2D grids are also probabilistic representation of the space 
and we assume the miniUGV’s position is known and unobserved grids have a uniform prior of being occupied. With the Bayesian filtering mechanism described in Sect.~\ref{sect:cam_model} and the {\em log odds} conversion~\cite{PRbook}, the probability that an individual grid $g_i$ is occupied given sensor measurements from time 1 to $t$ (i.e., $z_{1:t}$) can be obtained: 
\begin{equation}
    p(g_i|z_{1:t}) = \Bigg(1+\frac{1-p(g_i|z_t)}{p(g_i|z_t)} \frac{1-p(g_i|z_{1:t-1})}{p(g_i|z_{1:t-1})}\Bigg)^{-1}. 
\label{occ}
\end{equation}
Similarly, the grids are labeled free, occupied, or unknown based on the depth sensor input as $G = G_{free} \cup  G_{occ} \cup  G_{unk}$. 
We also define $F_g$ to be the global set of frontiers and $F_v$ to be the local set of frontiers limited to the fov of the IR sensors, such that $F_v \subseteq F_g$ as described in \cite{Cieslewski2017}.  

\subsection{Curiosity Heuristic for Hidden Object Exploration 
}

Philosopher and psychologist William James called curiosity ``the impulse towards better cognition", meaning that it is the desire to understand unknowns \cite{James_def}. We give the miniUGV a curiosity drive to know about the object hidden in the given space. This curiosity aims to answer two primary questions about the hidden space: does it contain the given object? And where is the object located?

To address this issue, we attach a curiosity value to each grid in $F_v$. 
According to the data from \cite{Kang2009}, decision-makers were least curious when they had no clue about the answer and also when they were extremely confident. They were most curious when they had some idea about the answer but lacked confidence. 
A study shows that the curiosity about the answer to a question is a U-shaped function of confidence about knowing that answer~\cite{Kidd2015}. 
And therefore, we define $c_i$ as the curiosity attached to $g_i, \forall \,\,  i \in N$, i.e., 
\begin{equation}
    \label{eq:curiosity}
    c_i= - \frac{(p(\omicron_i |\zeta_{1:t}, x_{1:t}) +a)^2}{4b} + \mathcal{K},
\end{equation}
where $a$, $b$ and $\mathcal{K}$ are constants determining the stiffness of the curiosity curve. These value must be set in a way that $c_i$ gets maximum value when $(p(\omicron_i |\zeta_{1:t}, x_{1:t}) \gets 0.5$, e.g., $a = -0.5, b = 0.1, \mathcal{K} = 0.62$ bounds $\mathcal{C}_i$ within $[0, 1]$.   

Remember that we have initialised $\mathcal{O}$ with a prior $p(\omicron_i) = 0.5$. This is an optimistic initialization as the miniUGV considers the probability of finding the object at each grid is $50\%$ prior to the start of a search operation. This initialization makes the miniUGV highly curious to look at each grid to find the given object. With each observation, miniUGV reduces its curiosity by finding that the object is either not present at a grid or the probability of the object's occupancy at $\omicron_i$ is higher than $0.5$ as considered in Eq. \eqref{cap_O}.


\begin{algorithm} [h!]
\label{alg:explore}
{\small
\KwResult{Explore the search space to find an object}
$t \gets time_{now}$\;
$a \gets -0.5$, 
$b \gets 0.1$ 
$\mathcal{K} \gets 0.62$\;
$G \gets 0.5$,
$O \gets 0.5$\;
Find $\mathcal{C}_t, F_g, F_v$\; 
\While{$F_v \neq \emptyset \And F_g \neq \emptyset$}
{
    Update $G, \mathcal{O}$\;
    Find $F_g, F_v$\;
    \eIf{$F_v \neq \emptyset $}
    {
        Find $\mathcal{C}_t$\;
        Find $p(\omicron_i |\zeta_{t+1}, x_{t+1})$~ $\forall~x_{t+1} \in F_v$\;
        Find $\mathcal{L(C)}$ $\forall F_v$\;
        Find $g*$, // using Eq. \eqref{eq:action_select}\;
        Execute $goto $ $g*$\;
    }
    {
        \eIf{$F_g \neq \emptyset $}
        {
            $g^n = nearest \, (g_i \in F_g$)\;
            Execute $goto$ $g^n$\;
        }
        {
            $\delta t = t - time_{now}$\;
            \textbf{return} $O, G, \delta t$\;
        }
    }
    }
}
\caption{Curiosity-Driven Hidden Space Exploration} 
\end{algorithm}
As the grids are independent, we assume the curiosity value at each grid is also independent of another and we measure the total curiosity attached to $\mathcal{O}$ at time $t$ as
\begin{equation}
    \mathcal{C}_t = \sum_{i=0}^{N} c_i = \sum_{i=0}^{N} - \frac{(p(\omicron_i |\zeta_{1:t}, x_{1:t}) +a)^2}{4b} + \mathcal{X}. 
\end{equation}
As mentioned in Sect. \ref{sect:cam_model}, the confidence in object detection varies with the distance of that object from camera, we estimate the observation value at time $t+1$ for miniUGV pose $x_{t+1}$ (pointing towards $\omicron_i$) as
\begin{equation}
    p(\omicron_i|\zeta_{t+1}, x_{t+1}) = \frac{p(\omicron_i|\zeta_t, x_t)d_t}{d_{t+1}},
\end{equation}
where $d_t$ and  $d_{t+1}$ are the distance of grid $\omicron_i$ from $x_t$ and $x_{t+1}$ respectively. 
Now, with this estimated observation, we get the $\mathcal{C}_{t+1}$ by applying equations Eq. \eqref{p_o_update}, Eq. \eqref{cap_O} and Eq. \eqref{eq:curiosity} sequentially. If the miniUGV performs an action to go to $x_i$, the expected curiosity loss $\mathcal{L(C)}$ can be calculated as 
\begin{equation}
\label{eq:C_loss}
    \mathcal{L(C)} = \mathcal{C}_t - \mathcal{C}_{t+1}. 
\end{equation}
Here, we only consider those grids which have updated values from camera measurements, 
as the other grids remain the same. 
The miniUGV always executes the action that can take it to the grid with the 
{highest curiosity loss} given by
\begin{equation}
\label{eq:action_select}
\begin{aligned}
   g_i^* = \operatorname*{argmax}_{g_i \in F_v}\, \{\mathcal{L(C)}\}\ \ 
    \textrm{s.t.}\, \, {p(o_i) > 0.5, \ \forall \omicron_i \in \mathcal{O}}. \\
\end{aligned}
\end{equation}
In essence, the mystery is uncovered (i.e., object discovered) when the curiosity becomes very low. 
{Alg. \ref{alg:explore} shows the 
exploration algorithm that attaches curiosity to the occupancy grid based on the confidence in object detection. Note that this curiosity can be added to any existing space exploration algorithms.}  


\section{Experimental Results}
\label{experimental_results}
The experimental setup consists of a DJI Matrice 100 UAV with DJI manifold onboard computer and DJI guidance kit for obstacle avoidance and vision-based localization. The UAV is equipped with a downward-facing Realsense D415 stereo vision camera and the miniUGV attached via the tethering mechanism. A major challenge faced in the setup was the communication between UAV and miniUGV system over WiFi, primarily because of the design of the miniUGV enclosure that made the signal reception a challenge. Also, since the magnetic encoders for the motors are mounted close to the compute board in the miniUGV, it affected the communication signal. This problem was tackled by limiting the communication bandwidth for only sharing the miniUGV position back to the UAV.
\begin{figure}[ht!] \vspace{-5pt}
{
\centering
    {
    \includegraphics[width=0.6\linewidth]{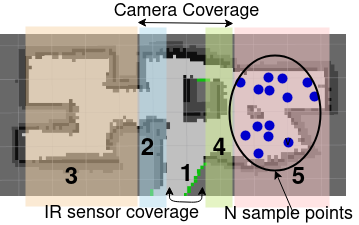}} \caption{Shows five different zones in the map to put the object for performance analyses. Blue points are the randomly selected position of the object in zone $5$ for evaluating the proposed algorithms.  
    } 
\label{fig:zone_map}
}
\end{figure}

\begin{figure*}[ht!] 
{
  \subfigure[Occupancy Grid Map1 ($G_{ir}$)]
  {\label{fig:G_map1}\includegraphics[height=1.in]{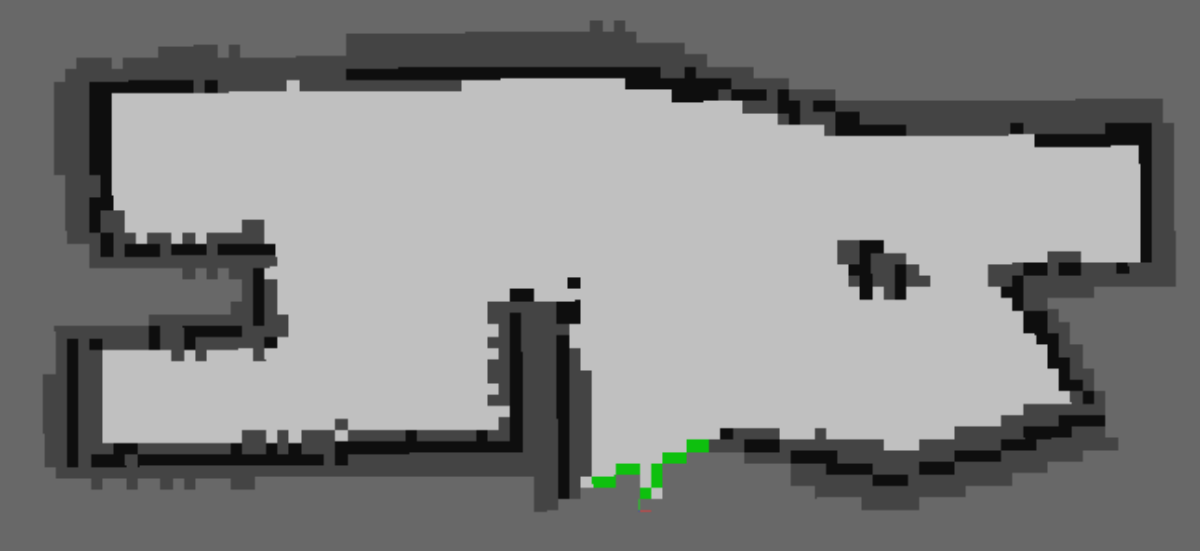}} \
  \subfigure[Object Map1 ($O_{cam}$)]
  {\label{fig:O_map1}\includegraphics[width=2.45in,height=1.in]{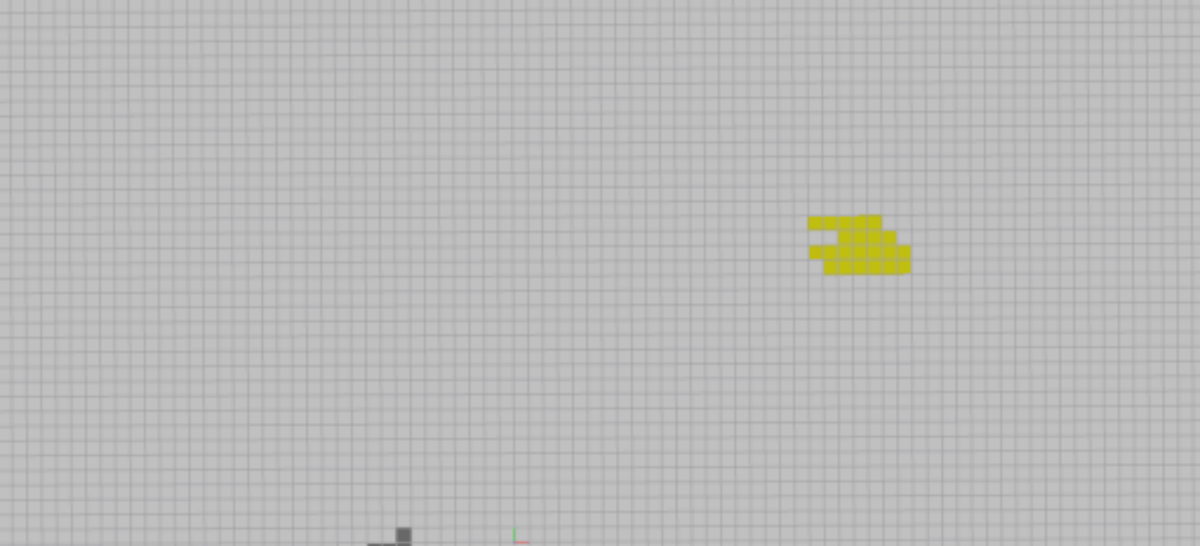}}
  \subfigure[$G$ and $O$ visualised togather]
  {\label{fig:G_O_map1}\includegraphics[height=1.in]{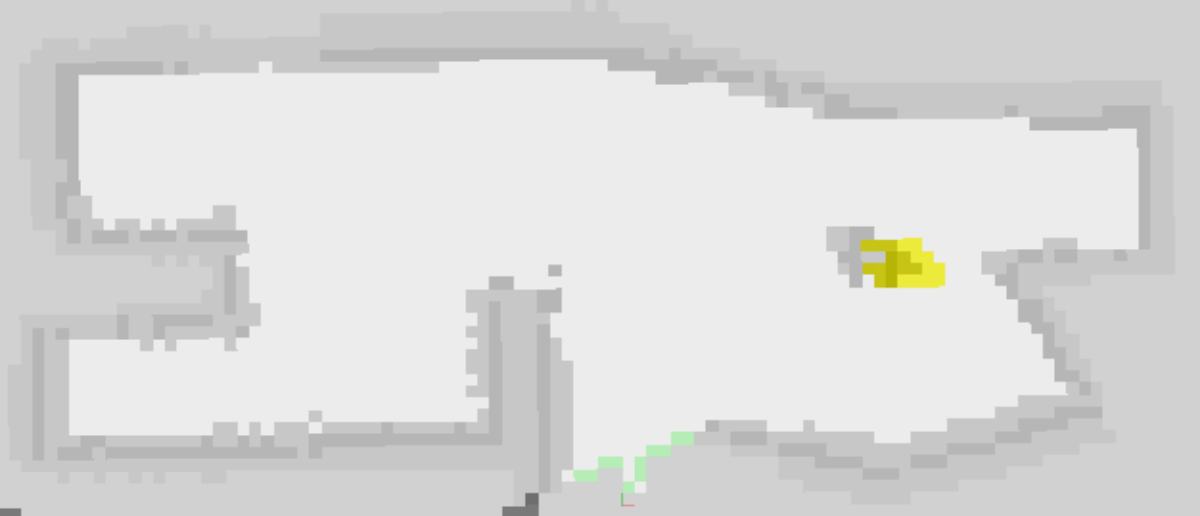} 
  }\vspace{-5pt}
  \subfigure[Occupancy Grid Map2 ($G_{ir}$)]
  {\label{fig:G_map2}\includegraphics[height=1.in]{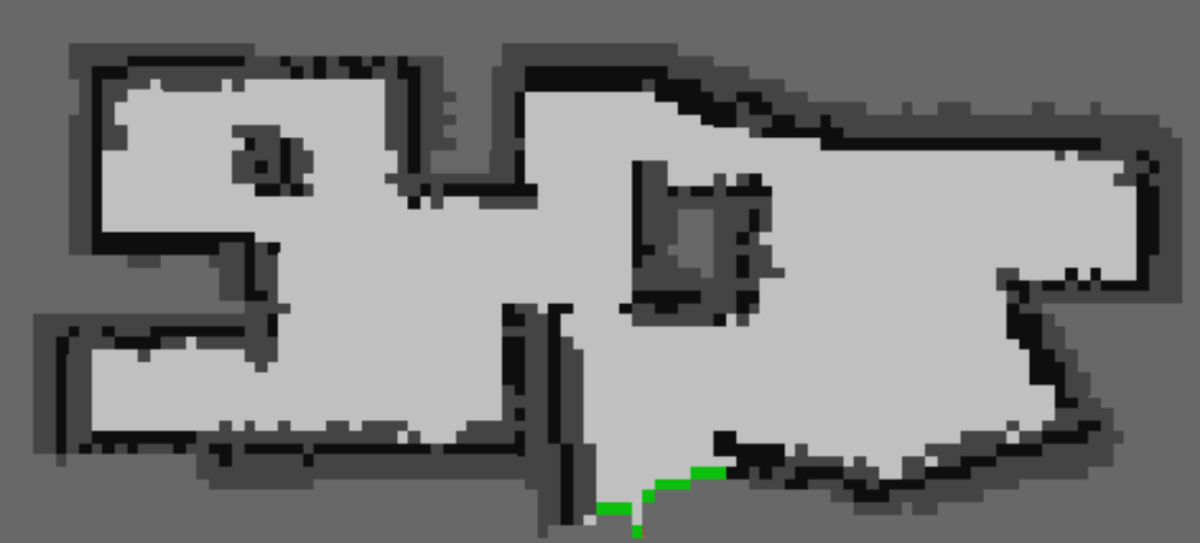}} \
  \subfigure[Object Map2 ($O_{cam}$)]
  {\label{fig:O_map2}\includegraphics[width=2.4 in, height=1.in]{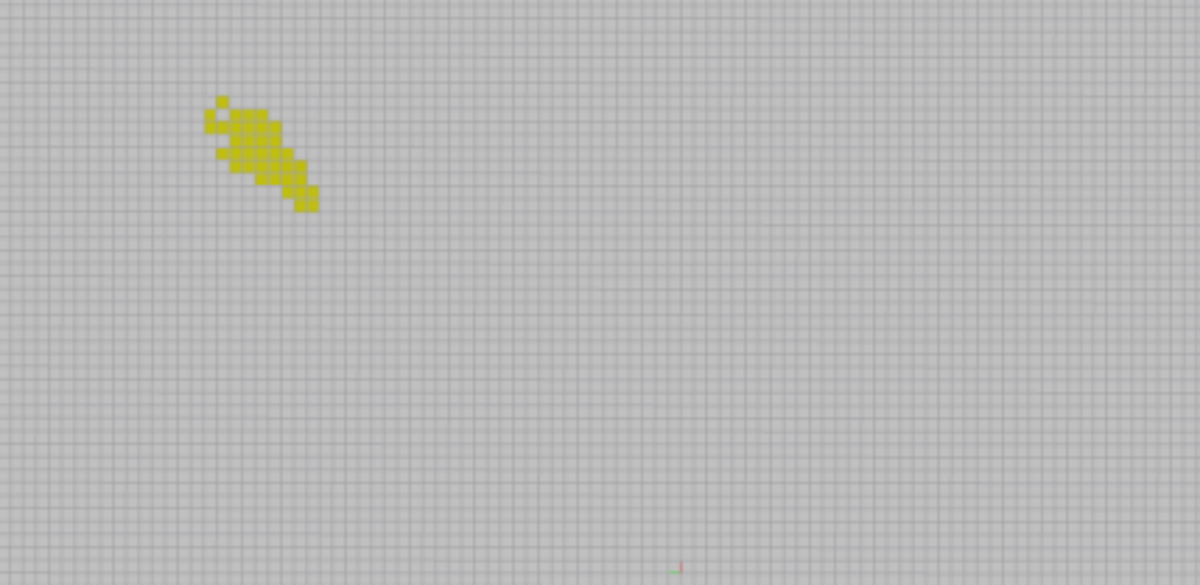}}
  \subfigure[$G$ and $O$ visualised togather]
  {\label{fig:G_O_map2}\includegraphics[height=1.in]{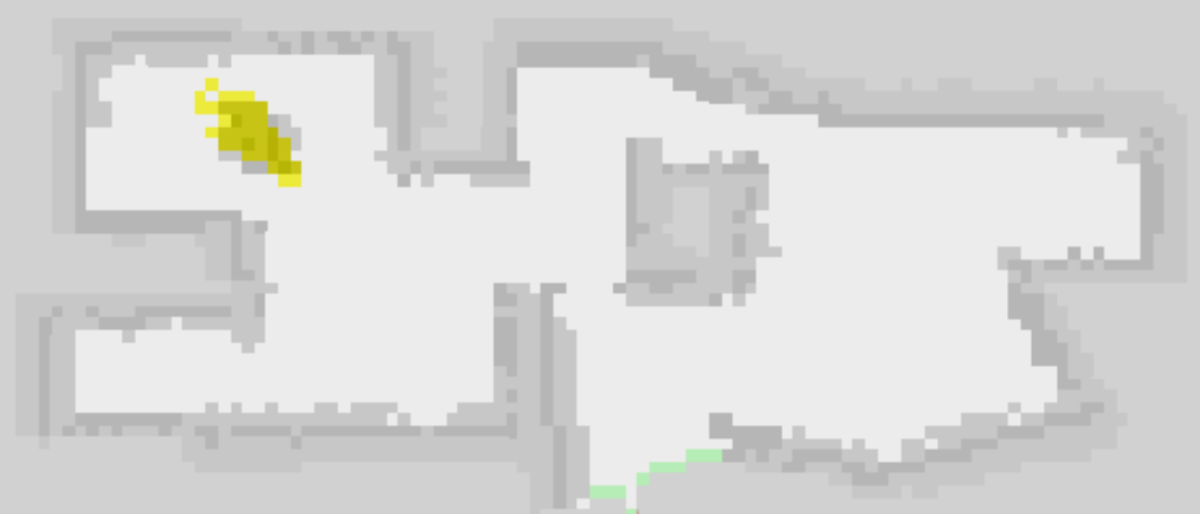}}  \vspace{-10pt}
\caption{Different maps generated after exploration. Yellow color shows the grids occupied with the object. Green color shows the frontiers.  \vspace{-15pt}}
\label{fig:two-maps} \vspace{-5pt}
}
\end{figure*}

\begin{figure}[t!] 
{
\centering
  \subfigure[]{\label{fig:exp_res1}\includegraphics[width=4.2cm, height=3.1cm]{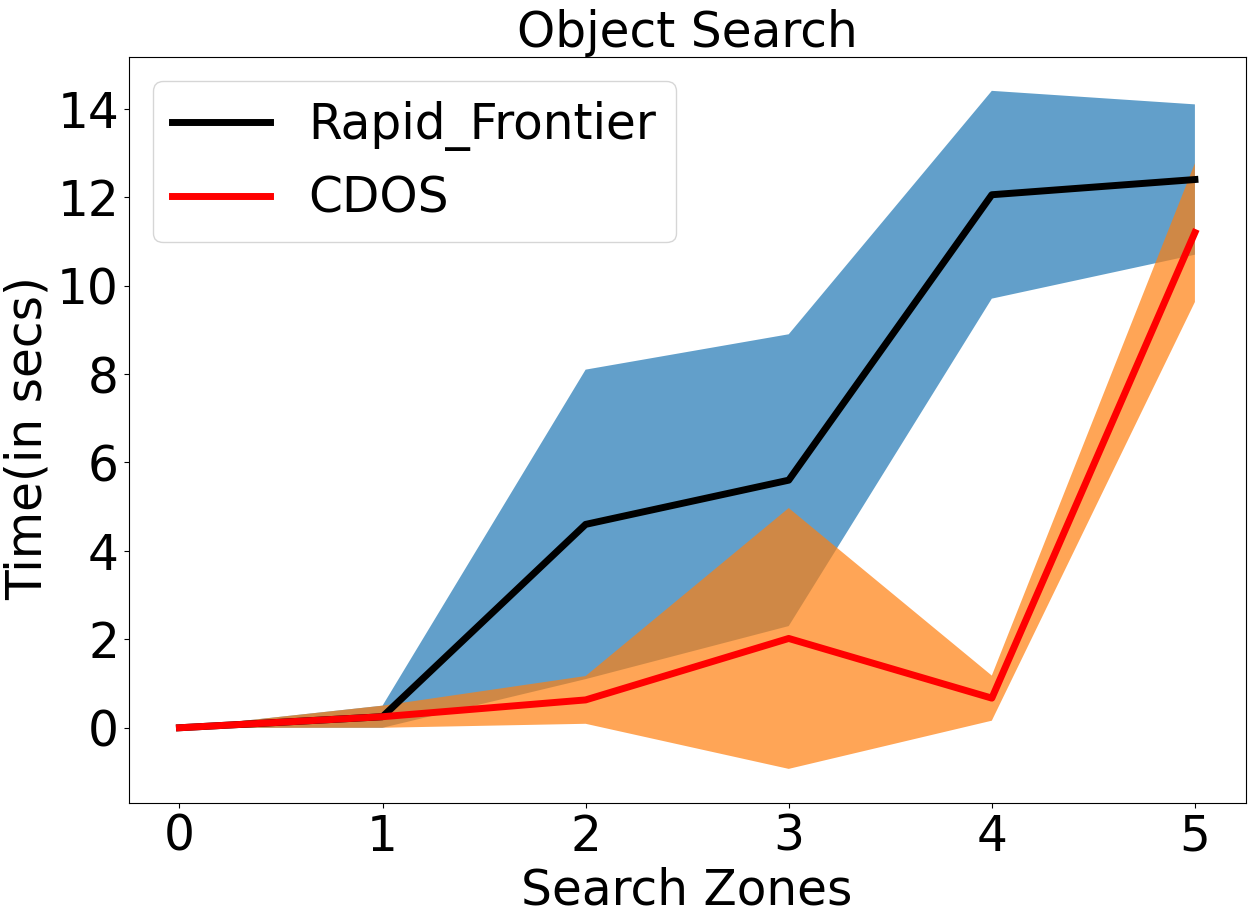}} 
  \subfigure[]{\label{fig:exp_res2}\includegraphics[width=4.2cm, height=3.1cm]{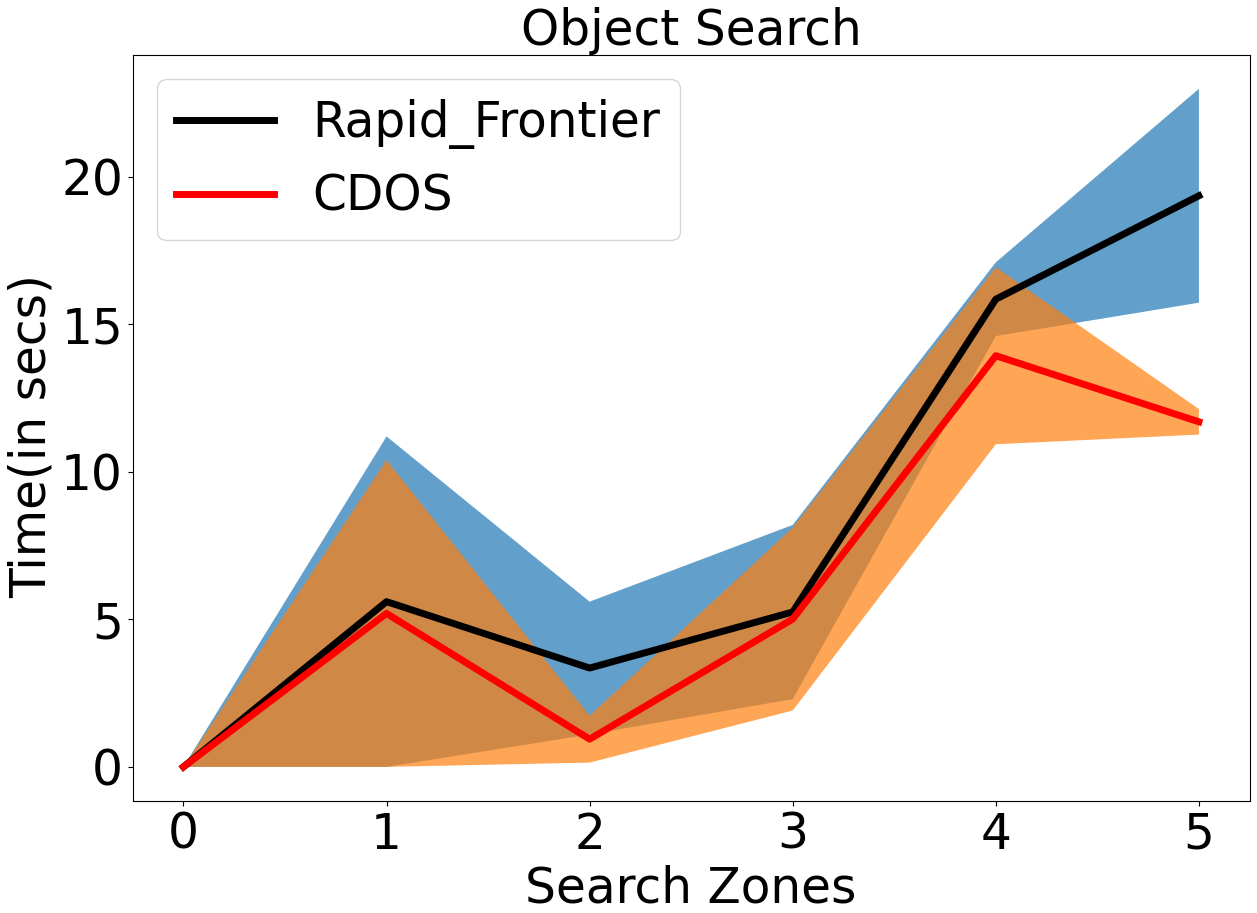}}  
  \caption{Performance evaluation on two different maps between the rapid frontier method and the proposed {curiosity driven object search (CDOS)} method. (a) Shows the performance evaluation on sparse map shown in Fig.~\ref{fig:G_map1}. (b) Shows the performance evaluation on dense map shown in Fig.~\ref{fig:G_map2}.
}
\label{fig:maps}
} 
\end{figure}

\begin{figure}[t!] 
{
\centering
  \subfigure[]{\label{fig:exp_fov_a}\includegraphics[width=4.1cm, height=2.6cm]{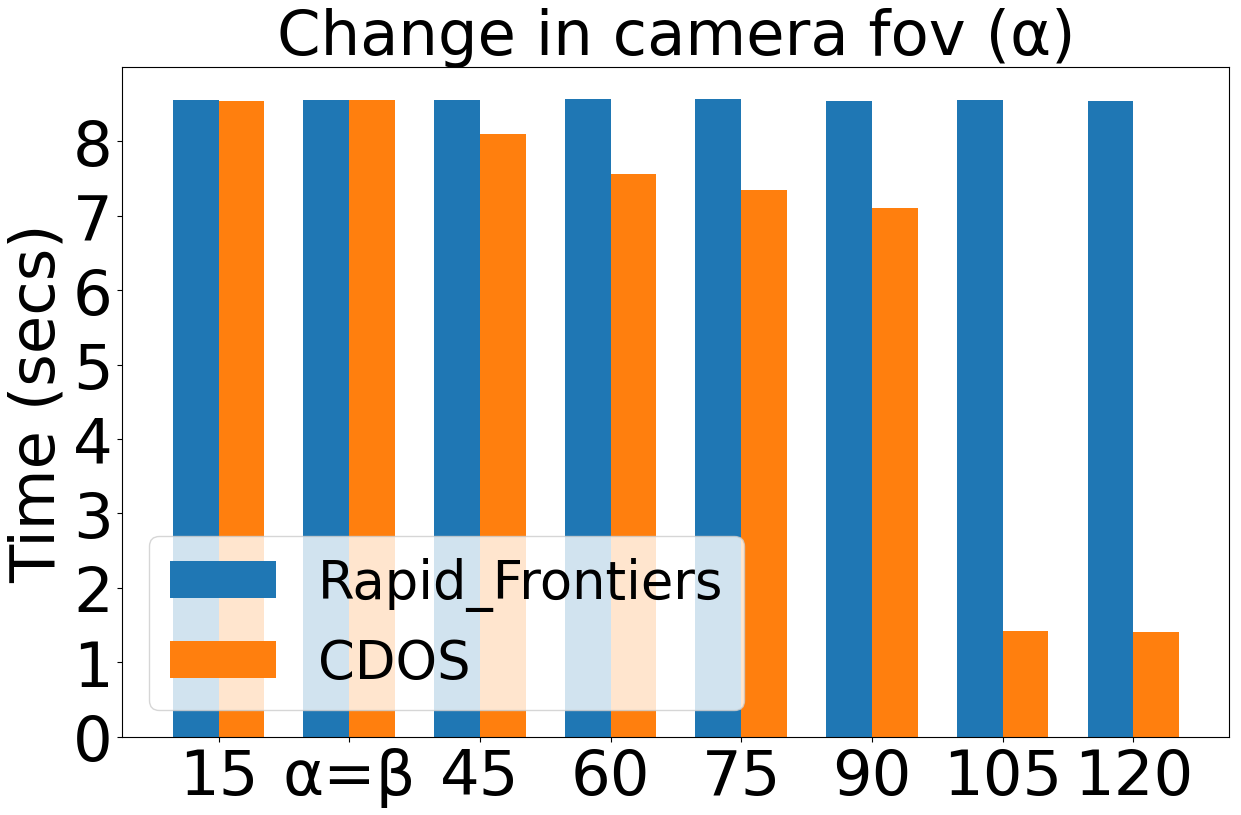}}\quad
  \subfigure[]{\label{fig:exp_fov_b}\includegraphics[width=4.1cm, height=2.6cm]{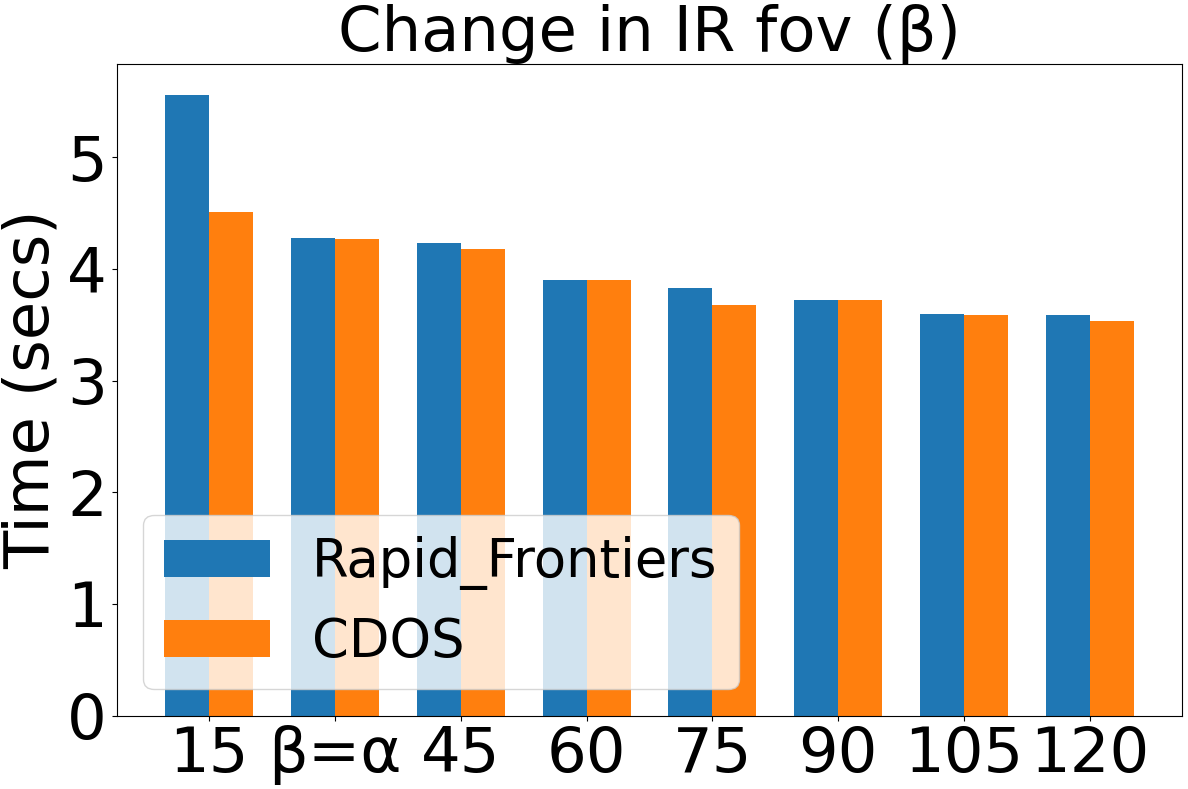}}
\caption{(a) Shows the performance evaluation by changing camera fov ($\alpha$). (b) Shows the performance evaluation by changing IR fov ($\beta$).  
}
\label{fig:maps1}
} \vspace{-15pt}
\end{figure}

\subsection{System Evaluation
}
All the system's functionalities are verified by conducting repetitive experiments in which the proposed system is given a task to eliminate an object (that might be a potential threat) from a hidden space as shown in Fig. \ref{fig:exp-images}. The UAV takes off and goes to the hidden space whose position is assumed to be known and detects a safe landing zone using the downward-facing depth camera. 
Fig. \ref{fig:exp-images} shows the detected landing zone during the autonomous mission. After landing, the system switches the mode of explorations from aerial to the ground and then the miniUGV performs curiosity-driven exploration for object search. The miniUGV creates two different types of map during exploration. Fig. \ref{fig:G_map1} and \ref{fig:G_map2} show the occupancy grid maps generated with the IR depth sensors and, Fig. \ref{fig:O_map1} and  \ref{fig:O_map2} show the object map created with the monocular camera modelled as a range finder. The final map is the combination of both as shown in Fig. \ref{fig:G_O_map1} and \ref{fig:G_O_map2}. To reduce the complexity of the mission, we used a simple scenario to evaluate the system performance in the real world. And therefore, the performance of the exploration algorithm has been rigorously tested in a simulated environment on two different maps shown in Fig. \ref{fig:G_map1} and \ref{fig:G_map2}.



\subsection{Exploration results}
We divide the map into five different zones as shown in Fig. \ref{fig:zone_map}. 
The miniUGV always starts exploration from zone $1$. Zone $2$ and $4$ are out of the IR sensors' reach and are only covered by the camera. Zone $3$ and $5$ are remaining spaces in the map. {Initially,} zone $2$ and $4$ are out of the IR sensors' reach and are only covered by the camera.
We sample $N$ points from each zone uniformly and place the object at each point one by one. 
We choose the state-of-the-art {\em rapid frontier exploration} \cite{Cieslewski2017} method as a comparison baseline. We repeat this process for both maps. 
{The baseline method uses camera to detect the object and it relies on IR sensors for space exploration. However, the proposed 
method is advantageous over the baseline since it uses the confidence in object detection while selecting the next frontier.}  

Results shown in Fig. \ref{fig:exp_res1} and \ref{fig:exp_res2} are generated with $max_{velocity}=2~m/s$, $D_{IR}=D_{cam}$ and $\alpha>\beta$. The target object is considered to be detected only when $conf>0.95$. Fig. \ref{fig:exp_res1} and \ref{fig:exp_res2} show the plots for average time taken to find the object in the maps shown in Fig. \ref{fig:G_map1} and \ref{fig:G_map2} respectively. Both the baseline method and our approach take the almost same time to find the object in Zone $1$ as the observations are available for both the sensors in Zone $1$. For all other zones, observations available for the camera are more than the IR sensors, and it is true for all $\alpha > \beta$. The curiosity values attached to each frontier drive the robots to deviate towards the object of interest even if the confidence in object detection is very low. As a result, the proposed algorithm always takes lesser time except for Zone $1$. 
Note that, the environment in Fig.~\ref{fig:G_map2} is more cluttered than in Fig.~\ref{fig:G_map1}. 
The chances of the object getting captured by the camera become lower in a more cluttered environment. If there is no detection (even with minimum probability), there would not be any curiosity to know about the object. Therefore, the performance gap between the two methods is low in a cluttered map.  

We then evaluated the proposed method on different values of $\alpha$ by keeping IR fov ($\beta$) constant at $30^o$. Fig. \ref{fig:exp_fov_a} clearly shows the proposed method is faster for all $\alpha > \beta$. Usually, the camera fov is larger than the IR fov (at least this is the case for miniUGV), these results show that considering the probability values obtained from object detection helps to find any object in lesser time than using existing algorithms for unknown area exploration by considering deterministic object detection. As a final comparison, we also tested our method for different values of $\beta$ and a constant $\alpha$. Fig. \ref{fig:exp_fov_b} shows that the time taken to find the abject using CDOS is at least equal to or less than the time taken by the rapid frontier method with deterministic object detection.

\section{Conclusion}
\label{conclusion}

In this work, we presented a novel UAV-miniUGV hybrid system for hidden area exploration. We have designed a miniUGV capable of autonomous exploration and tested its functionalities like object grabbing, visual servoing  navigation in several scenarios. The overall performance of the system has been tested by eliminating an object from a hidden space. We have designed and implemented a curiosity-driven object search heuristics, and validated its performance on various parameters. The proposed system takes advantage of a wider fov of the camera as well as the stochastic behavior of object detection. 
For future work, we would like to analyze the pros and cons of using a non-tethered pick and release of SR over a tethered counterpart discussed in this paper. Also, we plan to focus on the development of minimalistic hardware and software for lightweight exploration robots that can be dropped and retrieved by a UAV system. This is the first version and hence it has a huge potential to further improve. 
  


\bibliography{references} 
\bibliographystyle{ieeetr}

\end{document}